\documentclass[letterpaper]{article} 
\usepackage[preprint]{aaai2027}  
\usepackage[hyphens]{url}  
\usepackage{graphicx} 
\usepackage{natbib}  
\usepackage{caption} 
\usepackage[utf8]{inputenc}
\usepackage[T1]{fontenc}
\usepackage{microtype}
\usepackage{amsmath, amssymb, amsfonts}
\usepackage{booktabs}

\title{IFCLoRA: Topology-Aware Rank Allocation for Parameter-Efficient Fine-Tuning}
\author{Wei Zhang \quad Xinwu Liu \quad Yihang Cheng\corresponding}
\affiliations{Faculty of Artificial Intelligence in Education\\
Laboratory for Artificial Intelligence and New Forms of Education\\
Central China Normal University\\
Wuhan, Hubei, China}

\begin{document}

\maketitle

\begin{abstract}
Low-Rank Adaptation (LoRA) is a widely used approach to parameter-efficient fine-tuning (PEFT) of large language models, and its effectiveness depends on how ranks are allocated. Existing adaptive LoRA methods typically derive ranks from local gradient, activation, or matrix statistics collected either before or during fine-tuning. Training-time variants can add optimization overhead, while these local signals generally provide limited information about each module's structural role in task-relevant information propagation, offering little global grounding for allocating scarce adaptation capacity. To address this, we propose IFCLoRA (Information-Flow Centrality LoRA), a topology-aware method for pre-fine-tuning rank allocation and adapter initialization. Using a small calibration set, IFCLoRA performs lightweight intervention tracing on the frozen model and constructs a sparse task-conditioned interaction graph whose nodes are LoRA target modules. From this graph it extracts a global information-flow topology prior and fuses it with each node's local gradient sensitivity to form a topology-dominant Information-Flow Centrality (IFC) score, which measures how strongly each node participates in task-conditioned multi-hop propagation. The IFC scores then serve as module-level routing signals for one-shot discrete rank allocation under a total rank-budget constraint. Reusing the response vectors retained during tracing, IFCLoRA further constructs a function-preserving flow-response subspace initialization that gives adapters task-relevant initial output subspaces. Across all evaluated settings, IFCLoRA achieves higher mean scores than standard LoRA while keeping fine-tuning time and peak memory comparable; it additionally requires a one-time offline calibration stage. On mathematical reasoning (GSM8K), IFCLoRA attains the highest mean accuracy among the compared PEFT methods on both base models, exceeding standard LoRA by 4.75 percentage points on LLaMA-3.1-8B. The resulting rank allocations are non-uniform and vary across tasks and base models, suggesting that task-conditioned global information-flow topology can serve as a useful structural prior for rank allocation in low-budget PEFT.
\end{abstract}

\section{Introduction}

Full fine-tuning of large language models is costly, making parameter-efficient fine-tuning (PEFT) an attractive alternative \citep{han2024parameterefficient}. Among PEFT methods, LoRA represents weight updates with low-rank matrix products and is one of the most widely used baselines \citep{hu2022lora}.

Standard LoRA allocates ranks uniformly despite heterogeneous module roles. Existing adaptive methods improve over this baseline \citep{zhang2023adalora,he2025gora,ed-dib-etal-2025-gelora}, but they mainly rely on local activation, gradient, or matrix statistics and rarely model task-conditioned multi-hop interactions explicitly. Under a limited rank budget, module importance depends not only on local sensitivity but also on whether a module lies on task-relevant multi-hop propagation paths. Mechanistic tracing reveals task-relevant internal interactions \citep{conmy2023automated,syed2024attribution}, while graph centrality offers a structural view of node importance \citep{page1999pagerank,newman2018networks}. However, to our knowledge, task-conditioned interaction structure has not been explicitly converted into a pre-fine-tuning rank-allocation prior for LoRA. To close this gap, we propose IFCLoRA, a topology-aware method that combines intervention tracing with graph-based reachability for pre-fine-tuning rank allocation and adapter initialization.

\begin{figure}[t]
  \centering
  \includegraphics[width=0.90\linewidth]{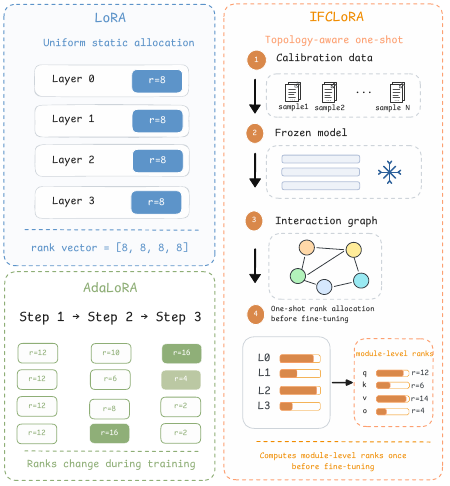}
  \caption{Rank allocation in LoRA (uniform), AdaLoRA (training-time dynamic), and IFCLoRA (topology-aware one-shot before fine-tuning).}
  \label{fig:lora-adalora-ifclora}
\end{figure}

Figure~\ref{fig:lora-adalora-ifclora} compares when and how LoRA, AdaLoRA, and IFCLoRA allocate adaptation capacity. IFCLoRA uses a shared pre-fine-tuning calibration stage to determine both how much rank each module receives and which output subspace its adapter initially occupies.

The main contributions of this paper are:
\begin{enumerate}
  \item \textbf{We revisit LoRA rank allocation from the perspective of task-conditioned information flow.} We identify the limitations of uniform allocation and training-time dynamic reallocation under tight budgets, and propose IFCLoRA, whose rank allocation and adapter initialization are both completed before fine-tuning begins, without introducing a dynamic rank-reallocation procedure during training.
  \item \textbf{We introduce Information-Flow Centrality (IFC) as a module-level routing signal.} IFC fuses the global topology captured by a task-conditioned interaction graph with local gradient sensitivity, providing a structured basis grounded in measured module interactions for allocating a limited rank budget. Reusing the response vectors retained by the same intervention-tracing pass, we further construct a function-preserving flow-response subspace initialization that supplies adapters with task-relevant initial output subspaces.
  \item \textbf{We evaluate IFCLoRA across multiple base models and tasks.} With total rank units matched, IFCLoRA outperforms standard LoRA and remains competitive with representative LoRA variants; rank-distribution analysis further reveals task- and model-dependent, non-uniform capacity routing.
\end{enumerate}

\section{Related Work}

\subsection{PEFT and Adaptive Rank Allocation}
\label{sec:related-peft}

Representative PEFT approaches include adapter-, prompt-, and low-rank parameterizations, as summarized by \citet{han2024parameterefficient}, with LoRA being a widely used low-rank baseline \citep{hu2022lora}. Adaptive-rank LoRA methods estimate importance during fine-tuning \citep{zhang2023adalora,liu2024alora,singh-etal-2025-l1ra}, or derive pre-fine-tuning signals from activations \citep{paischer2025eva,lin2026tlora}, gradients, and geometry \citep{ed-dib-etal-2025-gelora,he2025gora}. IGU-LoRA combines integrated gradients with uncertainty-aware scoring \citep{cui2026igulora,sundararajan2017axiomatic}, whereas DR-LoRA uses expert-routing frequency and gradient-based saliency for MoE adaptation \citep{deng2026drlora}. Data-driven initialization methods seed adapters from pretrained spectra \citep{meng2024pissa,buyukakyuz2024olora}, gradients \citep{wang2024loraga,he2025gora}, or activation variance \citep{paischer2025eva}. EVA uses activation explained variance for data-driven rank redistribution and initialization, whereas IFCLoRA converts intervention-traced, multi-hop reachability into a one-shot allocation prior that is lightly calibrated by local gradient sensitivity, and builds its initialization basis from the same traced responses without additional task-gradient information.

\subsection{Mechanistic Interpretability and Circuit Analysis}

Mechanistic interpretability seeks to understand how neural networks realize specific functions by examining internal computational structure.
Prior work suggests that language models contain traceable task-relevant computational subgraphs: attention heads, MLP layers, and residual pathways can jointly form circuits that implement specific functions \citep{elhage2021mathematical,wang2022interpretability,conmy2023automated,syed2024attribution}, while causal tracing and model editing locate localized computations that mediate factual recall \citep{meng2022locating}.
Studies of component roles further reveal the memory-like function of feed-forward networks \citep{geva2021transformer}.

Although these methods reveal important structural regularities, they mostly belong to the post hoc analysis paradigm.
The resulting graph structures have not yet been systematically used to guide parameter allocation before fine-tuning, and full circuit extraction is often computationally expensive.
IFCLoRA borrows the idea of modeling internal interactions as a directed graph, but redirects it from post hoc explanation to pre-fine-tuning rank allocation.

\subsection{Graph Centrality in Neural Networks}

Graph centrality is a classical tool in complex network analysis for measuring the relative importance of nodes in a network topology \citep{page1999pagerank,newman2018networks}.
A related line of work studies deep neural networks directly through sparsification and pruning.
Variational dropout induces sparsity in network weights \citep{molchanov2017variational}, whereas SNIP and the lottery-ticket hypothesis identify removable connections or trainable sparse subnetworks \citep{lee2019snip,frankle2019lottery}.
Centrality scores such as PageRank, Katz centrality, betweenness centrality, and eigenvector centrality motivate structural views of information flow for identifying important or redundant nodes.

However, existing graph-centrality methods mainly serve static architecture analysis or offline model compression.
For LLM adaptation, what affects downstream reasoning is not only whether a node is a static hub, but also whether it lies on a task-relevant path and acts as a mediator in multi-hop information propagation.
We therefore combine graph centrality with task-relevant local sensitivity to obtain the IFC metric introduced below.

Taken together, prior work shows that ranks should not be allocated uniformly and that downstream data, gradients, and attribution signals can provide useful allocation cues.
The gap we address is therefore not task awareness per se, but the explicit conversion of task-conditioned, multi-hop information-flow structure into a one-shot, budget-constrained rank-allocation prior before fine-tuning.

\begin{figure*}[t]
  \centering
  \includegraphics[width=0.98\textwidth]{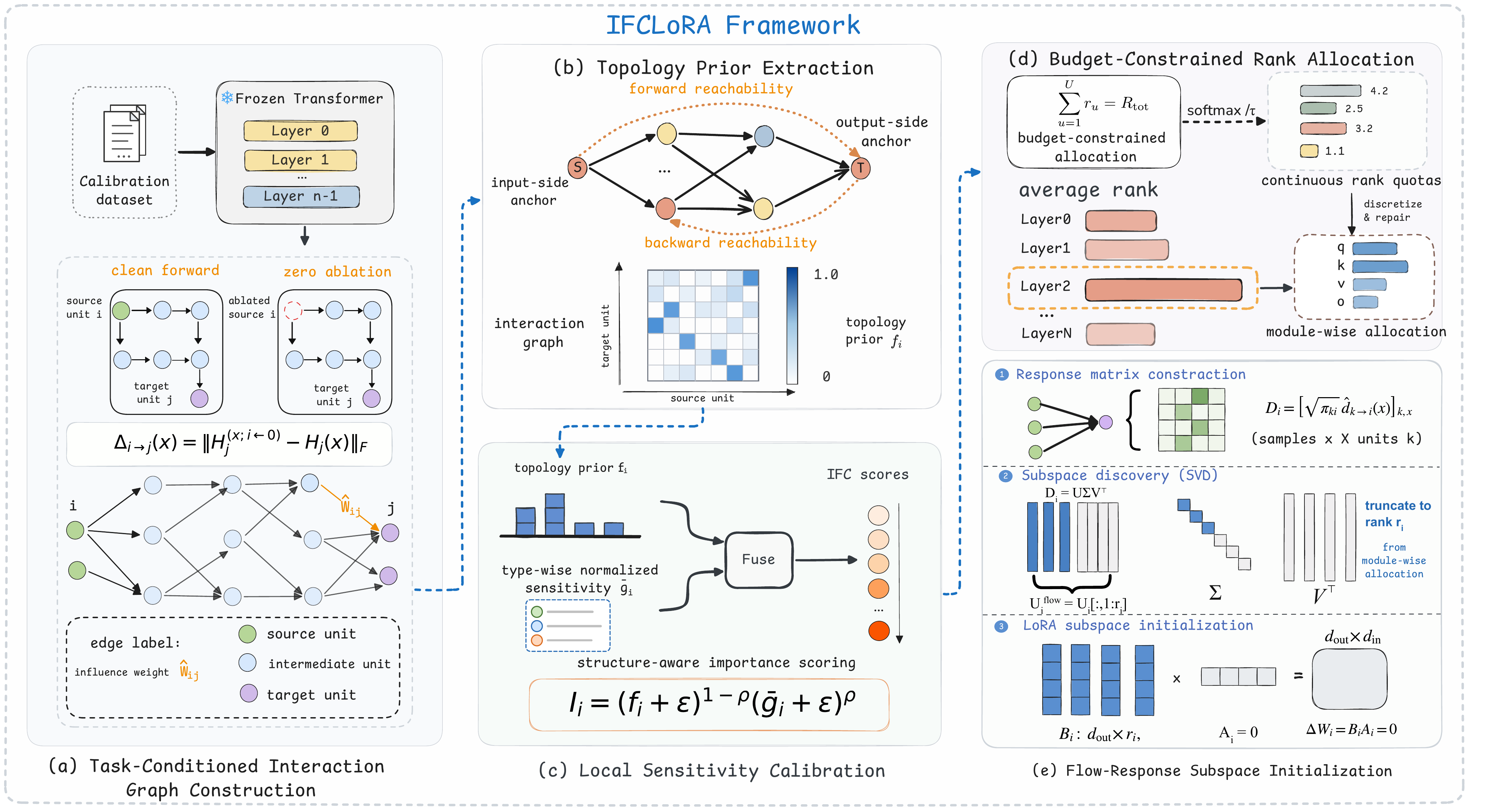}
  \caption{IFCLoRA workflow: (a) intervention tracing builds a task-conditioned graph; (b) reachability yields a topology prior; (c) fusion with normalized gradient sensitivity gives the IFC score; (d) budgeted allocation assigns module ranks; and (e) traced responses initialize function-preserving adapter output subspaces.}
  \label{fig:framework}
\end{figure*}

\section{Method}

We propose IFCLoRA, a topology-aware rank allocation method for low-budget PEFT. Unlike LoRA and training-time dynamic methods such as AdaLoRA \citep{hu2022lora,zhang2023adalora}, IFCLoRA completes allocation before fine-tuning begins, in four stages: task-conditioned interaction graph construction, topology-prior extraction, local sensitivity calibration, and budget-constrained rank allocation. A fifth stage, flow-response subspace initialization (FlowInit), reuses
the same tracing responses, so the overall calibration procedure
determines not only how much capacity each unit receives but also which
output subspace the corresponding adapter initially writes into.

Figure~\ref{fig:framework} illustrates the full workflow.

\subsection{Problem Definition}

Let the pretrained Transformer~\citep{vaswani2017attention} be $\mathcal{M}$, and let $\mathcal{U}=\{u_1,\ldots,u_{|\mathcal{U}|}\}$ denote all LoRA target modules, also referred to as adaptation units: the attention query, key, value, and output projection layers. Our goal is to assign an integer rank $r_u$ to each unit under a total rank budget $R_{\text{tot}}$, yielding $\mathbf{r}=(r_u)_{u\in\mathcal{U}}\in\mathbb{Z}^{|\mathcal{U}|}$:
\begin{equation}
\sum_{u\in\mathcal{U}} r_u = R_{\text{tot}}, \qquad
r_u \in \mathbb{Z}, \qquad
r_{\min} \leq r_u \leq r_{\max}.
\label{eq:budget}
\end{equation}
For the direct LoRA--IFCLoRA comparison, the matched quantity is total rank units rather than strictly identical trainable parameters, since target modules can have different shapes. The core problem is how to estimate module importance before training and use it for one-shot rank allocation.

\subsection{Task-Conditioned Interaction Graph Construction}

We represent interactions among LoRA target modules as a task-conditioned sparse directed interaction graph $G=(V,E)$.
Each node $v_i\in V$ corresponds to a LoRA target module, and each directed edge $(i\rightarrow j)\in E$ indicates that a change in node $i$ has an observable influence on a later node $j$ under the current task.
The goal is not to recover the full computation circuit exactly, but to build a sparse and tractable structural approximation that carries task-relevant information flow.

For each calibration sample $x\in\mathcal{D}_{\text{cal}}$, let $\mathbf{H}_j(x)\in\mathbb{R}^{T_x\times d_{\mathrm{out},j}}$ be node $j$'s token-level output over its $T_x$ non-padding tokens.
After zero-ablating source node $i$, we rerun the model to obtain $\mathbf{H}_j^{(x;i\leftarrow0)}$ at node $j$.
Zero ablation directly estimates how downstream representations change when a module contribution is removed, and avoids the extra activation-expectation estimates required by mean ablation.
The sample-level local influence strength is
\begin{equation}
\Delta_{i\rightarrow j}(x)=\left\|\mathbf{H}_j^{(x;i\leftarrow0)}-\mathbf{H}_j(x)\right\|_F,
\label{eq:delta}
\end{equation}
and the raw edge weight is
\begin{equation}
\widetilde{w}_{ij}=\mathbb{E}_{x\sim\mathcal{D}_{\text{cal}}}\left[\Delta_{i\rightarrow j}(x)\right].
\end{equation}
To control tracing cost, nodes are placed in a single linear order following the execution order in which adaptation units are visited during a forward pass, and candidate edges are retained only within a fixed window:
\begin{equation}
\widehat{w}_{ij}=\widetilde{w}_{ij}\cdot\mathbf{1}\!\left[i<j\leq\min(i+w,\,|\mathcal{U}|)\right],
\label{eq:filter}
\end{equation}
where $w=15$ in all experiments and $|\mathcal{U}|$ is the number of nodes. This linear index approximates the order in which modules interact; it is not a distance in the true computation graph. Within a block, the query, key, and value projections are computed in parallel, so their relative order is a traversal artifact, and it is the measured weight $\widetilde{w}_{ij}$ that ultimately determines whether a retained pair carries influence. We then row-normalize the retained weights, where $\mathcal{N}^{+}(i)=\{k\mid \widehat{w}_{ik}>0\}$ denotes the filtered out-neighborhood of node $i$:
\begin{equation}
T_{ij}=\frac{\widehat{w}_{ij}}{\sum_{k\in \mathcal{N}^{+}(i)}\widehat{w}_{ik}+\varepsilon}.
\end{equation}

\subsection{Topology Prior Extraction}

A node on a task-relevant information-propagation path should both receive task-relevant signals from the input side and pass influence toward the output decision side. We combine these properties with one input-side and one output-side anchor, represented by one-hot seed distributions $p^{\text{src}}$ and $p^{\text{tgt}}$:
\begin{equation}
\begin{aligned}
a &= \sum_{k=0}^{K_f}\alpha^k(T^\top)^k p^{\text{src}},\\
b &= \sum_{k=0}^{K_b}\beta^kT^k p^{\text{tgt}},\\
f_i &= \operatorname{Norm}(a_i\cdot b_i).
\end{aligned}
\end{equation}
Here $K_f$ and $K_b$ are truncation depths, $\alpha,\beta\in(0,1)$ are topology damping factors distinct from the LoRA scaling constant $\alpha_{\mathrm{LoRA}}$, and $\operatorname{Norm}$ denotes node-wise min-max normalization over all nodes. We use $K_f=K_b=15$ in all experiments. These propagation depths are distinct from the graph window $w=15$: $w$ limits the candidate target range of each source node, whereas $K_f$ and $K_b$ control the numbers of forward and backward multi-hop propagation steps, respectively. The resulting $f_i$ is a task-conditioned source-to-sink reachability score rather than an unconditional static centrality measure.

\subsection{Local Sensitivity Calibration and Budget-Constrained Rank Allocation}

The topology prior $f_i$ captures structural position, but capacity allocation should also reflect the supervised objective, so we lightly correct it with local gradient sensitivity on the same calibration set. For node $i$'s token-level output $\mathbf{H}_i(x)$,
\begin{equation}
g_i=\mathbb{E}_{x\sim\mathcal{D}_{\text{cal}}}\left[
\operatorname*{mean}_{t\in\mathcal{A}_x}
\left|\nabla_{\mathbf{H}_{i,t}(x)}\mathcal{L}(x)\odot\mathbf{H}_{i,t}(x)\right|_1\right],
\end{equation}
where $\mathcal{A}_x$ contains supervised answer tokens, the $\ell_1$ norm is over the hidden dimension, and the loss mask excludes prompt tokens.

Because gradient magnitudes differ systematically across module types, we
normalize $g_i$ within each type
$t(i)\in\{q,k,v,o\}$:
\begin{equation}
\bar{g}_i =
\frac{g_i}{
\frac{1}{|\mathcal{U}_{t(i)}|}
\sum_{j\in\mathcal{U}_{t(i)}} g_j + \varepsilon
}.
\end{equation}
We then fuse topology and sensitivity in power form:
\begin{equation}
I_i =
(f_i+\varepsilon)^{1-\rho}
(\bar{g}_i+\varepsilon)^{\rho}.
\end{equation}
Here $\rho\in[0,1]$ controls the balance between the topology prior and local sensitivity. For $0<\rho<1$, the multiplicative form acts as a soft AND gate: a node cannot obtain a high allocation score from either topology or gradient evidence alone, reducing the risk that a single noisy signal dominates allocation.

Since graph nodes and LoRA target modules correspond one-to-one, we set $S_u=I_u$ and apply z-score standardization over all units,
\begin{equation}
\widetilde{S}_u=\frac{S_u-\mu_S}{\sigma_S+\varepsilon},
\end{equation}
where $\mu_S$ and $\sigma_S$ are the mean and standard deviation over all LoRA target modules. A temperature-controlled softmax computes continuous rank quotas:
\begin{equation}
\omega_u=\frac{\exp(\widetilde{S}_u/\tau)}{\sum_{v\in\mathcal{U}}\exp(\widetilde{S}_v/\tau)}, \qquad
\widetilde{r}_u=R_{\text{tot}}\cdot\omega_u.
\end{equation}
The quota is clipped to $[r_{\min},r_{\max}]$ and floored; the residual budget is then resolved by a deterministic largest-remainder repair that adds one rank unit at a time to modules below $r_{\max}$, or removes one from modules above $r_{\min}$, breaking ties by module index. The repair satisfies Equation~\ref{eq:budget} exactly whenever $|\mathcal{U}|r_{\min}\leq R_{\mathrm{tot}}\leq|\mathcal{U}|r_{\max}$. All ranks are fixed before training; fine-tuning then follows standard LoRA with these non-uniform ranks and the initialization below. The window retains $O(|\mathcal{U}|w)$ candidate edges; the dominant preprocessing cost comes from clean and intervention forward passes over the calibration set.

\begin{table*}[t]
  \centering
  {\small
  \setlength{\tabcolsep}{1.5pt}
  \begin{tabular}{@{}lcccccc@{}}
    \toprule
    & \multicolumn{3}{c}{\textit{\mbox{LLaMA-3.1-8B}}} & \multicolumn{3}{c}{\textit{\mbox{LLaMA-2-7B}}} \\
    \cmidrule(lr){2-4}\cmidrule(lr){5-7}
    Method
    & \begin{tabular}[c]{@{}c@{}}MT-Bench\\Score $\uparrow$\end{tabular}
    & \begin{tabular}[c]{@{}c@{}}GSM8K Acc.\\ \textup{(\%)} $\uparrow$\end{tabular}
    & \begin{tabular}[c]{@{}c@{}}HumanEval\\pass@1 (\%) $\uparrow$\end{tabular}
    & \begin{tabular}[c]{@{}c@{}}MT-Bench\\Score $\uparrow$\end{tabular}
    & \begin{tabular}[c]{@{}c@{}}GSM8K Acc.\\ \textup{(\%)} $\uparrow$\end{tabular}
    & \begin{tabular}[c]{@{}c@{}}HumanEval\\pass@1 (\%) $\uparrow$\end{tabular} \\
    \midrule
    Full Fine-tuning & 5.91 $\pm$ 0.21 & 73.69 $\pm$ 0.35 & 51.42 $\pm$ 1.41 & 5.32 $\pm$ 0.12 & 59.19 $\pm$ 0.91 & 35.16 $\pm$ 1.96 \\
    LoRA & 6.13 $\pm$ 0.03 & 67.78 $\pm$ 1.25 & 42.89 $\pm$ 0.35 & 5.59 $\pm$ 0.09 & 42.08 $\pm$ 0.08 & 15.04 $\pm$ 0.35 \\
    rsLoRA & 6.20 $\pm$ 0.08 & 68.36 $\pm$ 0.74 & 45.53 $\pm$ 2.75 & 5.27 $\pm$ 0.04 & 45.49 $\pm$ 0.15 & 16.26 $\pm$ 0.70 \\
    DoRA & 6.22 $\pm$ 0.11 & 69.17 $\pm$ 1.00 & 44.11 $\pm$ 1.53 & \textbf{5.95 $\pm$ 0.03} & 53.22 $\pm$ 0.69 & 19.72 $\pm$ 0.35 \\
    LoRA+ & 6.36 $\pm$ 0.09 & 71.29 $\pm$ 0.93 & 44.92 $\pm$ 1.96 & 5.73 $\pm$ 0.07 & 51.96 $\pm$ 0.68 & 18.50 $\pm$ 0.70 \\
    OLoRA & 6.11 $\pm$ 0.05 & 68.54 $\pm$ 0.42 & 43.09 $\pm$ 2.31 & 5.32 $\pm$ 0.05 & 43.47 $\pm$ 0.79 & 17.07 $\pm$ 0.00 \\
    PiSSA & 6.10 $\pm$ 0.08 & 68.56 $\pm$ 1.03 & 44.31 $\pm$ 1.53 & 5.28 $\pm$ 0.03 & 44.38 $\pm$ 0.34 & 16.26 $\pm$ 0.35 \\
    LoRA-GA & 6.01 $\pm$ 0.05 & 71.39 $\pm$ 0.90 & 43.50 $\pm$ 0.70 & 5.92 $\pm$ 0.14 & 53.42 $\pm$ 0.37 & 20.12 $\pm$ 1.22 \\
    AdaLoRA & 6.17 $\pm$ 0.15 & 70.63 $\pm$ 0.77 & 41.87 $\pm$ 3.47 & 5.59 $\pm$ 0.06 & 50.72 $\pm$ 1.39 & 17.68 $\pm$ 0.61 \\
    GoRA & \textbf{6.37 $\pm$ 0.05} & 72.23 $\pm$ 0.72 & 48.37 $\pm$ 2.14 & 5.64 $\pm$ 0.11 & 54.18 $\pm$ 0.29 & \textbf{24.59 $\pm$ 1.27} \\
    \textbf{IFCLoRA} & 6.30 $\pm$ 0.06 & \textbf{72.53 $\pm$ 0.50} & \textbf{48.98 $\pm$ 1.76} & \textbf{5.95 $\pm$ 0.06} & \textbf{55.62 $\pm$ 0.48} & 24.39 $\pm$ 1.22 \\
    \bottomrule
  \end{tabular}
  }
  \caption{Main results at $r=8$ (mean $\pm$ standard deviation over three seeds). Bold marks the best PEFT mean per base model and metric; Full Fine-tuning is an unconstrained reference.}
  \label{tab:main-results}
\end{table*}

\subsection{Flow-Response Subspace Initialization}
\label{sec:flow-init}

The IFC score fixes how much capacity each unit receives, but not the directions in which that capacity is used. Graph construction uses all non-padding tokens to characterize sequence-level propagation, whereas local sensitivity and FlowInit restrict aggregation to supervised answer tokens to align with the fine-tuning objective. The tracing stage, however, already produces directional information that Equation~\ref{eq:delta} discards: before taking the Frobenius norm, the intervention response is a token-level matrix. For FlowInit, we mean-pool this response over supervised tokens to obtain a vector in the target unit's output space. For unit $u$ and an upstream source $k$ retained by Equation~\ref{eq:filter}, we use the supervised-token mean
\begin{equation}
d_{k\rightarrow u}(x)=\operatorname*{mean}_{t\in\mathcal{A}_x}
\left(\mathbf{H}_{u,t}^{(x;k\leftarrow0)}-\mathbf{H}_{u,t}(x)\right)
\ \in\ \mathbb{R}^{d_{\mathrm{out},u}},
\end{equation}
where $d_{\mathrm{out},u}$ is the output width. We index responses by their \emph{target} because $d_{k\rightarrow u}$ lies in $u$'s output space, matching the column space of $B_u$; outgoing responses $d_{u\rightarrow j}$ lie in other units' output spaces and are not used (see supplementary material).

Let $\mathcal{S}(u)=\{k:\widehat{w}_{ku}>0\}$ collect the retained upstream sources of $u$, weighted by $\pi_{ku}\propto(a_k\widehat{w}_{ku})^{\gamma}$ with $\sum_{k\in\mathcal{S}(u)}\pi_{ku}=1$, where $\gamma\in(0,1]$ tempers the influence of a few dominant edges. Stacking direction-normalized responses as columns over sources and calibration examples gives
\begin{equation}
\mathbf{D}_u=\Big[\sqrt{\pi_{ku}}\ \tfrac{d_{k\rightarrow u}(x)}{\|d_{k\rightarrow u}(x)\|_2+\varepsilon}\Big]_{k\in\mathcal{S}(u),\,x\in\mathcal{D}_{\text{cal}}},
\end{equation}
where $\mathbf{D}_u\in\mathbb{R}^{d_{\mathrm{out},u}\times |\mathcal{S}(u)||\mathcal{D}_{\mathrm{cal}}|}$. Because $\mathbf{D}_u\mathbf{D}_u^{\top}$ is a sum of outer products, it is positive semidefinite and invariant to sign flips of individual response columns. With $\mathbf{D}_u=\mathbf{U}_u\boldsymbol{\Sigma}_u\mathbf{V}_u^{\top}$, retaining the first $r_u$ left singular vectors $\mathbf{U}^{\mathrm{flow}}_u=[\mathbf{u}_{u,1},\ldots,\mathbf{u}_{u,r_u}]\in\mathbb{R}^{d_{\mathrm{out},u}\times r_u}$ selects the $r_u$ dominant response directions, where $r_u$ is the final integer rank after clipping, flooring, and deterministic budget repair.

We use these directions for a function-preserving initialization:
\begin{equation}
\begin{aligned}
B_u&=s_u\,\mathbf{U}^{\mathrm{flow}}_u,\qquad A_u=\mathbf{0},\qquad s_u=1/\sqrt{d_{\mathrm{out},u}},\\
\Delta W_u&=\frac{\alpha_{\mathrm{LoRA}}}{r_u}B_uA_u=\mathbf{0}.
\end{aligned}
\end{equation}
The initialized model therefore computes exactly the pretrained function. Because $A_u=\mathbf{0}$, the loss-gradient contribution at the first optimization step leaves $B_u$ at its initial value and updates $A_u$. Consequently, the induced effective weight update $B_u\Delta A_u$ lies in $\operatorname{col}(B_u)=\operatorname{span}(\mathbf{U}_u^{\mathrm{flow}})$ (see supplementary material for the gradient derivation); optimizer regularization may separately change a nonzero $B_u$ when applied. Thus the graph selects the output subspace the adapter first writes into while the task gradient supplies the matching input-side directions. Degenerate units---those with no retained incoming edge, whose total unnormalized FlowInit source weight is at or below $\epsilon_{\mathrm{resp}}$, or whose $\mathbf{D}_u$ has fewer nonzero singular values than $r_u$---fall back to standard LoRA initialization; the fallback conditions and computational accounting are reported in the supplementary material. Conditional on the allocated ranks, FlowInit constructs its output basis entirely from traced responses and graph weights, without requiring additional task-gradient information. This distinguishes its basis construction from weight- or gradient-based initialization schemes such as PiSSA, OLoRA, LoRA-GA, and GoRA.

\section{Experiments}

At $r=8$, we cover mathematical reasoning, instruction following, and code generation, each with three seeds; the $r=16$ experiments cover GSM8K only.

\subsection{Experimental Setup}
\label{sec:experimental_setup}

We fine-tune the base checkpoints of \mbox{LLaMA-3.1-8B} and \mbox{LLaMA-2-7B} \citep{dubey2024llama3,touvron2023llama2} in three independently trained settings: MetaMathQA $\rightarrow$ GSM8K \citep{yu2023metamath,cobbe2021gsm8k}, WizardLM $\rightarrow$ MT-Bench \citep{xu2023wizardlm,zheng2023mtbench}, and Code-Feedback $\rightarrow$ HumanEval \citep{zheng2024opencodeinterpreter,chen2021humaneval}. Metrics are GSM8K exact-match after answer extraction, HumanEval pass@1, and MT-Bench scores obtained with the judge protocol specified in the supplementary material. Compared methods are Full Fine-tuning, LoRA \citep{hu2022lora}, rsLoRA \citep{kalajdzievski2023rslora}, DoRA \citep{liu2024dora}, LoRA+ \citep{hayou2024loraplus}, OLoRA \citep{buyukakyuz2024olora}, PiSSA \citep{meng2024pissa}, LoRA-GA \citep{wang2024loraga}, AdaLoRA \citep{zhang2023adalora}, GoRA \citep{he2025gora}, and IFCLoRA. All PEFT methods are applied to the $q,k,v,o$ projections and use seeds 42, 52, and 62.

We study nominal budgets $r=8$ and $r=16$. For nominal budget $r$, the total rank-unit budget is $R_{\text{tot}}=r|\mathcal{U}|$, whereas the realized trainable adapter count is
\begin{equation}
P_{\mathrm{adapter}}
=
\sum_{u\in\mathcal U}
r_u\left(d_{\mathrm{in},u}+d_{\mathrm{out},u}\right).
\label{eq:adapter-parameters}
\end{equation}
Because the $q,k,v,o$ projections can have different input and output widths, one rank unit need not have the same parameter cost across modules. Direct LoRA--IFCLoRA comparisons therefore match total rank units, while adaptive baselines retain their native budget mechanisms; exact realized counts are reported in the supplementary material.

IFCLoRA uses a 128-example calibration set with $\rho=0.25$, $\tau=0.8$, rank bounds $[2,32]$, window $w=15$, and damping factors $\alpha=\beta=0.85$. Unless otherwise stated, IFCLoRA denotes IFC-based non-uniform rank allocation with FlowInit. Dataset preprocessing, optimization schedules, generation settings, answer extraction, anchor definitions, numerical stabilizers, LoRA scaling, and FlowInit fallback rules are fully specified in the supplementary material. Results are reported as mean $\pm$ standard deviation over three training seeds $\{42,52,62\}$.

\paragraph{Hyperparameter selection.}
Unless varied in an ablation, IFCLoRA hyperparameters are fixed once using the protocol reported in the supplementary material and then shared across model-task settings; test sets are used only for final evaluation under that protocol.

\subsection{Main Results}
\label{sec:main-results}

Table~\ref{tab:main-results} reports the main results at $r=8$. IFCLoRA achieves the highest mean GSM8K accuracy among the evaluated PEFT methods on both base models, improving over LoRA by 4.75 percentage points on \mbox{LLaMA-3.1-8B} and 13.54 percentage points on \mbox{LLaMA-2-7B}. It also achieves the highest mean HumanEval pass@1 on \mbox{LLaMA-3.1-8B} and remains competitive on the remaining HumanEval and MT-Bench settings.

The following analyses examine whether these gains persist across rank budgets, which components produce them, and which alternative explanations the controls can rule out.

\subsection{Rank Budget Scaling}
\label{sec:rank-budget-scaling}

\begin{table}[t]
  \centering
  {\small
  \setlength{\tabcolsep}{3pt}
  \begin{tabular}{@{}llcc@{}}
    \toprule
    Base Model & Method & $r=16$ & $\Delta$ from $r=8$ \\
    \midrule
    \mbox{LLaMA-3.1-8B} & LoRA & 69.12 $\pm$ 0.95 & +1.34 \\
     & AdaLoRA & 71.57 $\pm$ 0.69 & +0.94 \\
     & IFCLoRA & \textbf{73.54 $\pm$ 0.50} & +1.01 \\
    \midrule
    \mbox{LLaMA-2-7B} & LoRA & 44.66 $\pm$ 0.80 & +2.58 \\
     & AdaLoRA & 52.31 $\pm$ 1.10 & +1.59 \\
     & IFCLoRA & \textbf{57.67 $\pm$ 0.55} & +2.05 \\
    \bottomrule
  \end{tabular}
  }
  \caption{GSM8K accuracy (\%) at $r=16$ and change in mean from $r=8$ over three seeds. The $r=8$ results are reported in Table~\ref{tab:main-results}; bold marks the best $r=16$ mean.}
  \label{tab:gsm8k-budget}
\end{table}

Table~\ref{tab:gsm8k-budget} shows performance across two rank budgets. At $r=16$, IFCLoRA remains ahead of LoRA and AdaLoRA on both base models, so its advantage persists at both evaluated budgets under the protocol defined in Experimental Setup. We next separate the contributions of allocation and initialization.

\subsection{Ablation Study}
\label{sec:ablation-study}

\begin{figure*}[t]
    \centering
    \includegraphics[width=\textwidth]{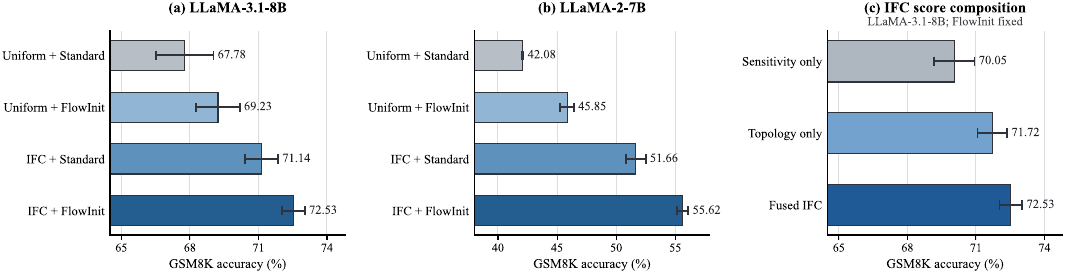}
    \caption{GSM8K ablations at $r=8$ (mean $\pm$ standard deviation over three seeds). (a,b) Component ablations on \mbox{LLaMA-3.1-8B} and \mbox{LLaMA-2-7B} isolate IFC rank allocation and FlowInit. (c) Score-composition ablation on \mbox{LLaMA-3.1-8B} with FlowInit fixed compares sensitivity-only, topology-only, and fused IFC.}
    \label{fig:ablation_combined}
\end{figure*}

\paragraph{Component-level ablation.}
Figure~\ref{fig:ablation_combined}(a,b) separates the effects of IFC-based rank allocation and FlowInit. On both base models, replacing uniform allocation with IFC allocation improves performance under standard initialization, while FlowInit also improves performance under uniform ranks. Their combination achieves the highest mean accuracy.

\paragraph{IFC score composition.}
Figure~\ref{fig:ablation_combined}(c) further ablates the IFC score composition on \mbox{LLaMA-3.1-8B} while holding FlowInit fixed. Topology-only allocation outperforms sensitivity-only allocation, and the fused IFC score yields the highest mean accuracy, suggesting that topology provides the stronger standalone allocation prior while local sensitivity offers a complementary signal.

Having established the component contributions, we next use matched controls to test alternative explanations for the gains.

\subsection{Mechanism Analysis}
\label{sec:mechanism-analysis}

\paragraph{Topology permutation control.}
On both base models, shuffled IFC preserves the realized rank multiset and total rank units while permuting only the correspondence between ranks and module identities. Under the budget protocol in Experimental Setup, the original placement achieves higher means than the shuffled placement: $71.14 \pm 0.72$ versus $68.84 \pm 0.91$ on \mbox{LLaMA-3.1-8B}, and $51.66 \pm 0.83$ versus $46.37 \pm 0.88$ on \mbox{LLaMA-2-7B}. This supports the claim that module-specific placement matters; exact parameter counts after permutation are reported in the supplementary material.

\paragraph{Scaling control.}
This control is reported only on \mbox{LLaMA-3.1-8B}. Holding the IFC rank vector and standard initialization fixed, replacing $\alpha_{\mathrm{LoRA}}/r_u$ with $\alpha_{\mathrm{LoRA}}/r_{\mathrm{nominal}}$ changes GSM8K accuracy from $71.14 \pm 0.72$ to $70.74 \pm 0.79$. The latter remains 2.96 percentage points above Uniform LoRA, so rank-dependent scaling alone does not account for the allocation gain.

\begin{figure}[t]
    \centering
    \includegraphics[width=\columnwidth]{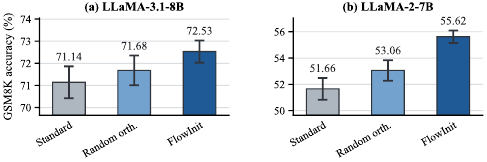}
    \caption{Initialization-direction controls on GSM8K at $r=8$ for (a) \mbox{LLaMA-3.1-8B} and (b) \mbox{LLaMA-2-7B}. IFC ranks and the scaling policy are fixed; bars show protocol-specific means and standard deviations.}
    \label{fig:initialization-controls}
\end{figure}

\paragraph{Initialization direction control.}
Figure~\ref{fig:initialization-controls} fixes IFC ranks and scaling on both base models. Random orthogonal bases control for generic orthogonal parameterization, while FlowInit uses task-conditioned response directions. FlowInit exceeds the random control by 0.85 and 2.56 percentage points on \mbox{LLaMA-3.1-8B} and \mbox{LLaMA-2-7B}, respectively, suggesting that the traced directions carry information beyond generic orthogonality; held-out-gradient analyses are reported in the supplementary material.

The quantitative controls isolate placement and initialization effects; we next examine the resulting allocation patterns descriptively.

\subsection{Qualitative Analysis}
\label{sec:qualitative-analysis}

\begin{figure}[t]
    \centering
    \includegraphics[width=\linewidth]{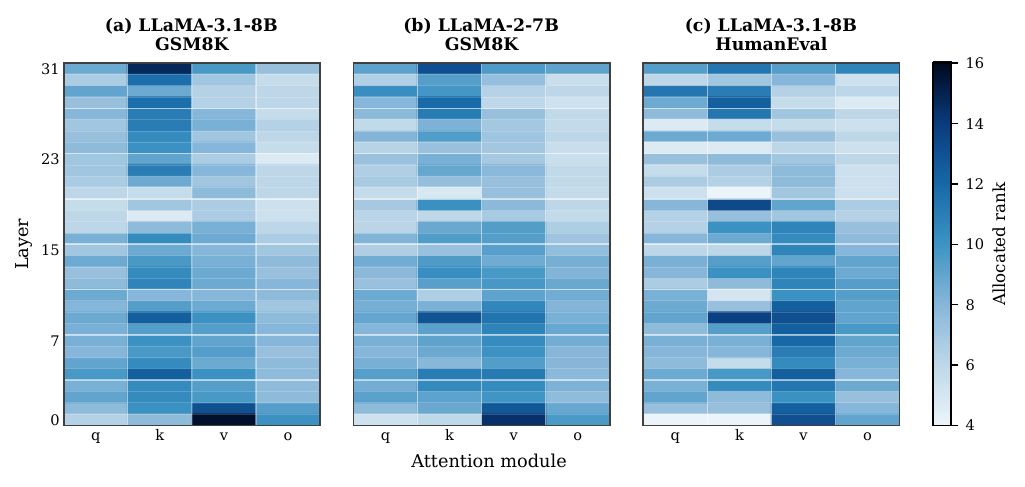}
    \caption{IFCLoRA ranks across attention projections and layers for (a) \mbox{LLaMA-3.1-8B} on GSM8K, (b) \mbox{LLaMA-2-7B} on GSM8K, and (c) \mbox{LLaMA-3.1-8B} on HumanEval.}
    \label{fig:rank-allocation}
\end{figure}

Figure~\ref{fig:rank-allocation} shows non-uniform layer- and projection-specific profiles. The two GSM8K base models share some trends but differ in detail, while GSM8K and HumanEval produce distinct profiles on the same base model. These descriptive patterns are consistent with task-conditioned rather than fixed architecture-level allocation; rank-vector statistics are reported in the supplementary material.

Finally, we examine initialization stability and computational cost.

\subsection{Stability and Efficiency}
\label{sec:stability-efficiency}

\paragraph{Initialization stability.}
\label{sec:initialization-variance}
This control is reported only on \mbox{LLaMA-3.1-8B}. Resampling a random orthogonal basis for each training seed yields $71.49 \pm 0.93$, while reusing one fixed random basis yields $71.68 \pm 0.67$, a 0.19-point mean difference. With the same IFC ranks and scaling, FlowInit reaches $72.53 \pm 0.50$. This suggests that, across the three seeds, fixed-basis reuse alone does not explain the FlowInit gain; full seed-level results are reported in the supplementary material.

\paragraph{Resource accounting.}
On \mbox{LLaMA-3.1-8B} GSM8K at $r=8$ under the same A800 hardware and training protocol, LoRA and IFCLoRA require 3h29m and 3h32m for fine-tuning, with peak GPU memory of 44.0 GB and 44.3 GB, respectively. IFCLoRA additionally requires a one-time 34-minute offline calibration stage. Intervention passes, response-cache size, decomposition cost, fallback frequency, and full wall-clock accounting are reported in the supplementary material.

\section{Discussion and Limitations}
IFC is a heuristic without formal optimality guarantees and provides structural evidence, not a complete causal account. FlowInit likewise treats intervention sensitivity as a proxy for optimization usefulness: traced response directions need not be those that best reduce task loss. The graph covers $q,k,v,o$ projections but not FFN modules, so it represents attention-projection interactions rather than the complete Transformer computation graph. Its unnormalized token-level Frobenius edge norm (Equation~\ref{eq:delta}) may depend on output dimensionality. Matching total rank units does not guarantee matched parameter counts across heterogeneous projections; we report realized counts and leave exact parameter matching to future work.

The current method uses one input and one output anchor; multi-anchor variants remain untested. Random-orthogonal and held-out-gradient controls provide directional evidence, but several mechanism comparisons use only three seeds. Slightly larger first-step update-to-gradient ratios under FlowInit leave update magnitude as a possible secondary factor. Component attribution is evaluated only on GSM8K. Transfer across capabilities, domains, larger or newer models, encoder-decoder and mixture-of-experts architectures, broader NLU benchmarks, and other insertion locations remains to be established.
\section{Conclusion and Future Work}
IFCLoRA combines intervention-derived topology and local sensitivity for one-shot rank allocation, then reuses traced responses for function-preserving initialization. Matched controls provide evidence of benefits beyond generic orthogonal initialization. Future work includes cheaper or adaptive graph construction, stronger causal validation, and extension to broader models, modalities, contexts, and PEFT settings.

\bibliography{references}

\end{document}